 \documentclass[tablecaption=bottom,wcp]{jmlr} 

\usepackage{booktabs}
\usepackage[load-configurations=version-1]{siunitx} 
\usepackage{algorithm}
\usepackage{algorithmic}

\usepackage{xcolor}         
\newcommand{\ap}[1]{{\color{orange}[AP: #1]}}

\newcommand{\LL}[1]{{\color{red}[LL: #1]}}

\theorembodyfont{\upshape}
\theoremheaderfont{\scshape}
\theorempostheader{:}
\theoremsep{\newline}

\jmlrproceedings{Preprint}{Preprint}

\title[Individual Fairness in Bayesian Neural Networks]{Individual Fairness in Bayesian Neural Networks}

\author{\Name{Anonymous Authors}\\
  \addr Anonymous Institution}

\author{
\Name{Alice Doherty} \addr Trinity College Dublin \Email{adohert@tcd.ie} \\
\Name{Matthew Wicker} \addr The Alan Turing Institute \Email{m.wicker@turing.ac.uk} \\
\Name{Luca Laurenti} \addr Delft University of Technology \Email{l.laurenti@tudelft.nl} \\
\Name{Andrea Patane} \addr Trinity College Dublin \Email{apatane@tcd.ie} \\
}

\begin{document}

\maketitle

\begin{abstract}
We study Individual Fairness (IF) for Bayesian neural networks (BNNs).
Specifically, we consider the $\epsilon$-$\delta$-individual fairness notion, which requires that, for any pair of input points that are $\epsilon$-similar according to a given similarity metrics, the output of the BNN is within a given tolerance $\delta>0.$
We leverage bounds on statistical sampling over the input space and the relationship between adversarial robustness and individual fairness to derive a framework for the systematic estimation of $\epsilon$-$\delta$-IF, designing \textit{Fair-FGSM} and  \textit{Fair-PGD} as global, fairness-aware extensions to gradient-based attacks for BNNs.  
We empirically study IF of a variety of approximately inferred BNNs with different architectures on fairness benchmarks, and compare against deterministic models learnt using frequentist techniques. Interestingly, we find that BNNs trained by means of approximate Bayesian inference consistently tend to be markedly more individually fair than their deterministic counterparts. 



\end{abstract}

\section{Introduction}
\label{sec:intro}

Deep learning models have achieved state-of-the-art performance in a wide variety of tasks \citep{goodfellow2016deep}.
Several calls for caution have, however, recently been raised about their deployment in tasks where fairness is of concern \citep{barocas2016big}. In fact, Neural Networks (NNs) have been found to reinforce negative biases from sensitive datasets \citep{bolukbasi2016man}, discriminating against individuals on the basis of attributes such as gender or race.
To address this, research efforts have been directed at both measuring the fairness of NNs, their de-biased training, as well as defining precise notions of fairness.
%

 Given a BNN $f$ and a similarity metric between individuals $d_{fair}$, which encodes a task-dependent notion of similarity \citep{ilvento2019metric},  \textit{Individual Fairness} (IF) enforces that all pairs of \textit{similar} individuals in the input space get treated similarly by $f$ \citep{dwork2012fairness}.  As opposed to the statistical nature of \emph{group fairness} \citep{mehrabi2021survey},  IF aims at computing worst-case bias measures over a model input space.   
Consequently, albeit being defined on the full input space and over a fairness similarity metric, because of its worst-case nature, IF has been linked to adversarial robustness \citep{yeom2020individual}. 
As it has recently been show that Bayesian Neural Networks (BNNs) have a tendency to be less fragile to adversarial attacks than their frequentist counter-parts \citep{carbone2020robustness}, it is natural to wonder whether approximate Bayesian inference may also have a positive impact over the IF of a neural network. 
However, to the best of our knowledge, no work has been conducted along these lines of inquire.

In this paper, we investigate the IF of BNNs and empirically evaluate it on various benchmarks. 
While exact computations of IF in BNNs is infeasible due to their non-convexity, we exploit the relationship between IF and adversarial robustness \citep{yeom2020individual,benussi2022individual}  to develop a framework for the adaptation of adversarial attack methods for IF.
In particular, we explicitly instantiate \textit{Fair-FGSM} and \textit{Fair-PGD} as extensions of their corresponding adversarial attacks \citep{goodfellow2014explaining,athalye2018obfuscated} by employing gradient steps modifications and projections specific to $d_{fair}$ metrics commonly used in the fairness literature.
Furthermore, while attack methods estimate IF locally  around a given individual, we use concentration inequalities \citep{boucheron2004concentration} to statistically bound the worst-case expected IF over all pairs of individuals.  

We perform an empirical evaluation of IF in BNNs on a selection of benchmarks, including the Adult dataset \citep{UCIdatsets} and Folktables \citep{ding2021retiring}, and on a variety of different architectural, approximate Bayesian inference, and similarity metric parameters.
We compare the results obtained by BNNs with those obtained by deterministic NNs and deep ensembles. 
We find that BNNs consistently outperform their deterministic counterparts in terms of individual fairness. That is, albeit still learning certain biases, all things being equal,  BNNs show a tendency to be \textit{fairer} than their deterministic counter-parts. 
We finish the paper with an empirical analysis  inquiring this interesting property and a discussion on why approximate Bayesian inference may lead to fairer prediction models.

\section{Individual Fairness for BNNs}\label{sec:fairness}
We consider a NN $f^w : X \rightarrow \mathbb{R}^m$ trained on a dataset  $\mathcal{D} = \{ (x_i,y_i)\}_{i=1}^{N} $ composed by $N$ points, where $w$ is the aggregated vector of weights and biases, $X \subset \mathbb{R}^n$ is the input space, and $n$ and $m$ are respectively the input and output dimensions.
For simplicity, we focus on the case of binary classification, however, our results can be trivially extended to the regression and multi-class case.
In a Bayesian setting, $w$ is sampled from a random variable $\mathbf{w}$ whose posterior distribution is obtained by application the Bayes rule:
$ p(w|\mathcal{D}) \propto p(\mathcal{D}|w)p(w) $,
where  $p(w)$ is the prior and $p(\mathcal{D}|w)$ is the likelihood.
The posterior predictive distribution $\pi$ over an input $x$ is then obtained  by averaging the posterior distribution over a sigmoid function $\sigma$, i.e.,  
$\pi(x) = \int \sigma(f^w(x))p(w|\mathcal{D})dw .$ 
To define individual fairness for BNNs, we adopt the $\epsilon$-$\delta$-IF \citep{john2020verifying}, which depends on a similarity metric $d_{fair}: X  \times X \rightarrow \mathbb{R}_{\geq 0 }  $  encapsulating task-dependent information about sensitive attributes. 
\begin{definition}\label{def:if}
Given $\epsilon \geq 0$ and $\delta \geq 0$,  we say that a BNN $f^{\mathbf{w}}$ is $\epsilon$-$\delta$-individually fair iff $
    \forall x', x'' \in X s.t.\ d_{fair}(x',x'') \leq \epsilon  \implies \vert \pi(x') - \pi(x'') \vert \leq \delta,$ 
that is, if the predictive distribution of $\epsilon$-similar individuals are $\delta$-close to each other.
\end{definition}
Notice that while related to notion of adversarial robustness for BNNs \citep{cardelli2019statistical, wicker2020probabilistic}, Definition \ref{def:if} has some key differences.
First, rather than being a local definition specific to a test point and its neighbourhood, individual fairness looks at the worst-case behaviour over the whole input space, i.e., simultaneously looking at neighbourhoods around all the input points in $X$.
Furthermore, while adversarial attacks are generally developed for an $\ell_p$ metric, individual fairness is built around a task-specific similarity metric $d_{fair}$. 
Intuitively, $d_{fair}$ needs to encode the desired notion of similarity between individuals \citep{dwork2012fairness}. 
While a number of metrics have been discussed in the literature \citep{ilvento2019metric}, we focus on the following ones, which are widely used and
can be automatically learnt from data \citep{yurochkin2019training}:\footnote{While not directly investigated in this paper, our fairness attacks can be generalised to similarity metrics built over embeddings \citep{ruoss2020learning}, by attacking the embeddings as well.}
\begin{itemize}
\item \textbf{Weighted $\mathbf{\ell_p}$.} In this case $d_{\text{fair}}(x',x'')$ is defined as a weighted version of an $\ell_p$ metric, i.e.\
$d_{\text{fair}}(x',x'') = \sqrt[p]{\sum_{i=1}^n \theta_i |x'_i - x''_i|^p}  $, where the  weights $\theta_i$ can be set accordingly to their correlation with the sensitive attribute.
\item \textbf{Mahalanobis distance.} In this case we have $d_{\text{fair}}(x',x'')  = \sqrt{(x'-x'')^T S^{-1} (x'-x'')}$, for a given positive semi-definite (SPD) matrix $S$. Intuitively, $S$ accounts for the intra-correlation of features to capture latent dependencies w.r.t.\ the sensitive features.
\end{itemize}
\section{Fairness attacks}\label{sec:attacks}
Definition \ref{def:if} could be reformulated according to the following optimization problem
\begin{align}\label{eq:optimisation}
    \delta^* = \max_{x' \in X} {\max_{\substack{x'' \in X\\ d_{fair}(x',x'') \leq \epsilon  }} \vert \pi(x') - \pi(x'') \vert}.
\end{align}
 Checking IF is then equivalent to check whether $\delta^* \leq \delta $ or not.
In Eqn \eqref{eq:optimisation}, the inner maximization problem finds the point in a specific $d_{fair}$-neighbourhood that maximizes the variation in the predictive posterior, while the outer one considers the global part of the definition.
We proceed by solving the inner-optimisation problem with gradient-based techniques, and rely on statistical methods to solve with high confidence the outer problem.
\paragraph{Inner Problem}
Because of the non-convexity of NN architectures, exact computation of the inner maximization problem in Eqn \eqref{eq:optimisation} is generally infeasible.
We proceed by adapting gradient-based optimization techniques
commonly used to detect adversarial attacks to our setting. In particular, in the case of the Fast Gradient Sign Method (FGSM) \citep{goodfellow2014explaining}, given the log-loss $L$ associated to the likelihood function (e.g., the binary entropy for the sigmoid), and a point $x$ with associated label $y$, we obtain that:
$$ x_{x',\max} \approx x' +\eta \cdot  \text{sgn}( \mathbb{E}_{w \sim p(w|\mathcal{D})}[\nabla_x L(  f^w(x),y  )  ]).$$
$\eta \in \mathbb{R}^n$ scales the attack strength according to  $\epsilon$ and $d_{fair}$. It is the first ingredient that needs to be adapted for $\epsilon$-$\delta$-IF.
In the case of the weighted $\ell_p$ metric, $\eta$ can be simply defined as the vector with entries $\eta_i = \epsilon/\sqrt{\theta_i}$. 
For the Mahalanobis distance, given $x'$ and $\epsilon$, $d_{fair}(x',x'') \leq \epsilon$ describes a hyper-ellipsoid centred in $x'$.
We can hence set $\eta$ to be the vector whose generic component is scaled according to the ellipsoid axis, i.e., $\eta_i = \epsilon \sqrt{S_{ii}}$.
Furthermore, while one is guaranteed to remain inside the $d_{fair}$-neighbourhood in the $\ell_p$ case, for the Mahalanobis metric one needs to check whether the attack is inside the $\epsilon$-$d_{fair}$ ball or not. In the latter case, we set up a local line search problem to project $x_{x',\max}$ back to a point inside the ball. 
Other gradient-based attacks such as PGD \citep{madry2017towards} or CW \citep{carlini2017magnet} can be similarly adapted by taking consecutive gradients. Similarly, other more general attacks could be employed \citep{yuan2020gradient,daubenerefficient} as well as training techniques \citep{liu2018adv, wicker2021bayesian}. In Section \ref{sec:Results}, we will consider both FGSM and PGD, which we denote with \emph{Fair-FGSM} and \emph{Fair-PGD} to distinguish them from the adversarial attack counter-parts. 




\paragraph{Outer Problem}
The final step needed for the computation $\epsilon$-$\delta$-IF is to check whether: 
\begin{align}\label{eq:verification}
\max_{x' \in X}\big( \pi(x_{x',\max})- \pi(x') \big)  \leq \delta.
\end{align}
Again, exact computation of Eqn \eqref{eq:verification} is infeasible as it would require solving a non-convex optimisation problem over a possibly large input space. We relax the problem and instead compute the probability that a point sapled from $X$ is fair. In order to do that we iid sample $n_{\text{samples}}$ points from $X$ and build the following empirical estimator:
$\hat{p}=\sum_{i=1}^{n_{\text{samples}}} \frac{\phi(x_i)}{n_{\text{samples}}}, \quad \text{where }  \phi(x_i)=\begin{cases} 1 &\text{if }\pi(x_{x_i,\max}) - \pi(x_i)  \leq \delta \\
0 & \text{otherwise}
\end{cases}.$ 
Note that $\phi(x_i)$ is a binary variable. Consequently, $\hat{p}$ approximate the probability that a point sampled from $X$ satisfies the inner condition in Eqn \eqref{eq:optimisation}. Lemma \ref{proposition:Chernoff} below is a straightforward application of the Chernoff bound \citep{hellman1970probability} and guarantees that if $n_{\text{samples}}$ is large enough, then the approximation error introduced by the empirical estimator can be made arbitrarily small with high confidence.
\begin{lemma}
\label{proposition:Chernoff}
Assume that $n_{\text{samples}}$ points are sampled iid from $X$ according to a (possibly unknown) probability measure $P.$ 
Call $ p=\mathbb{E}_{x'\sim P}[ \phi(x')]. $
Then for any $\gamma,\theta >0$ such that  $ n_{samples} > \frac{1}{2\theta^2} \log \left( \frac{2}{\gamma} \right),$ it holds that
$ P( |\hat{p} - p| > \theta\big) \leq \gamma. $
\end{lemma}
Therefore, in order to approximate the solution of the optimisation problem in Equation \ref{eq:verification} with statistical confidences $\theta$ and $\gamma$  we compute the smallest $\delta$ such that $\hat{p}=1$ with $n_{samples}$ points randomly sampled from the input space.

\section{Results}
\label{sec:Results}
\begin{figure}[h]
    \centering
        \includegraphics[trim={0 0 3.7cm 0},clip,width=0.29\columnwidth]{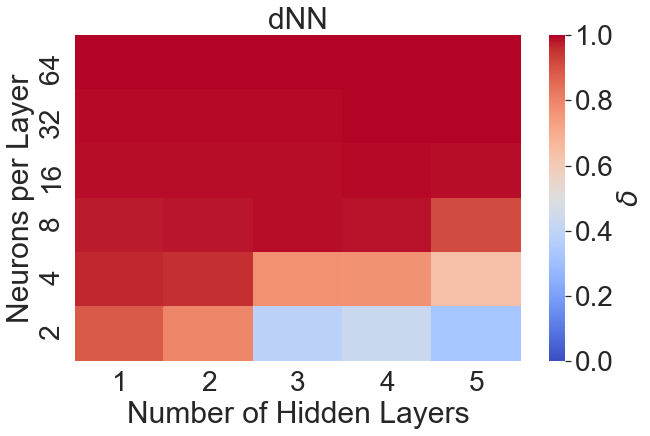}
    \includegraphics[trim={0 0 3.7cm 0},clip, width=0.29\columnwidth]{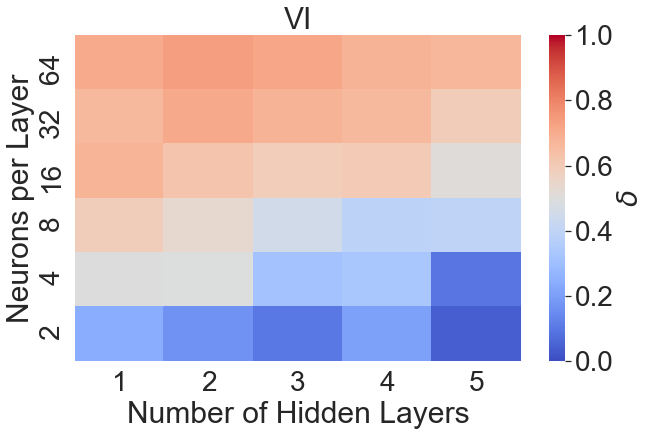}
    \includegraphics[width=0.348\columnwidth]{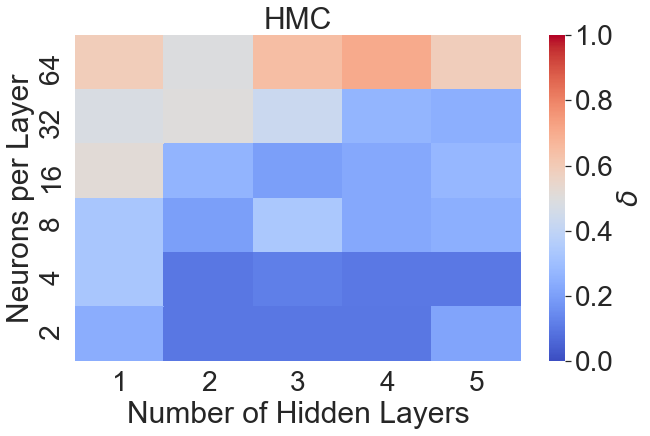}
        \includegraphics[trim={0 0 3.7cm 0},clip, width=0.29\columnwidth]{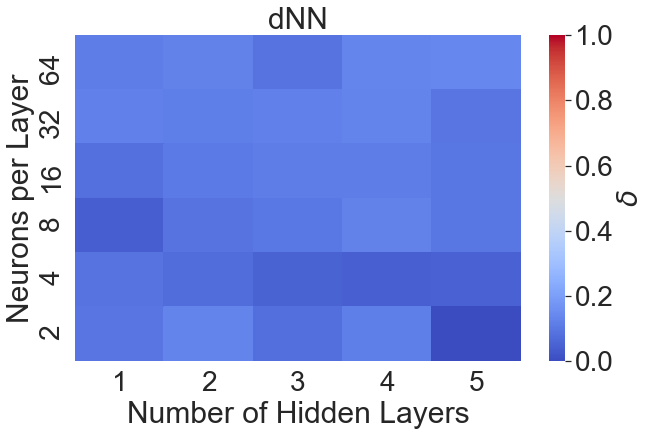}
    \includegraphics[trim={0 0 3.7cm 0},clip, width=0.29\columnwidth]{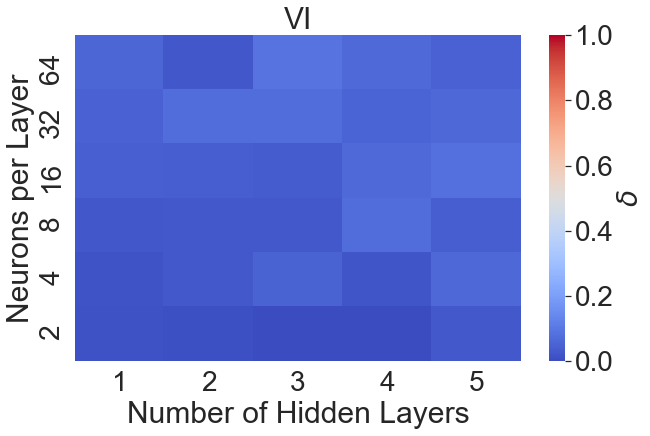}
    \includegraphics[width=0.348\columnwidth]{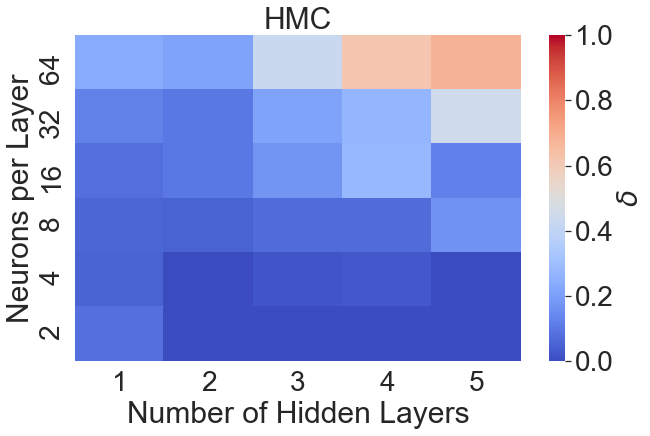}
    \includegraphics[trim={0 0 3.7cm 0},clip, width=0.29\columnwidth]{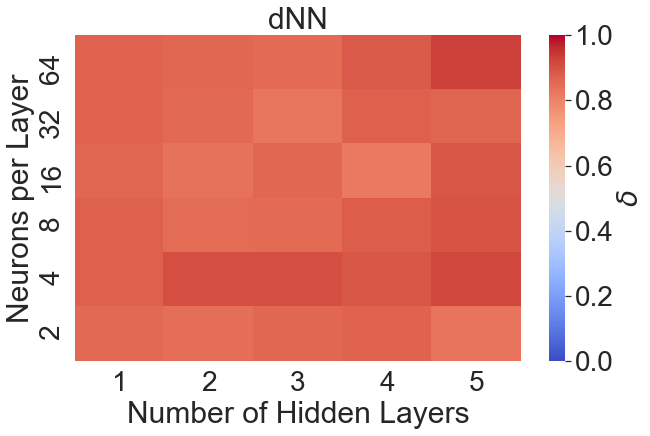}
    \includegraphics[trim={0 0 3.7cm 0},clip, width=0.29\columnwidth]{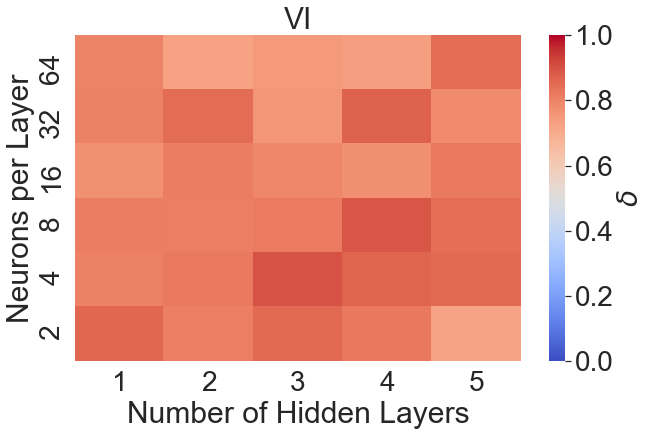}
    \includegraphics[width=0.348\columnwidth]{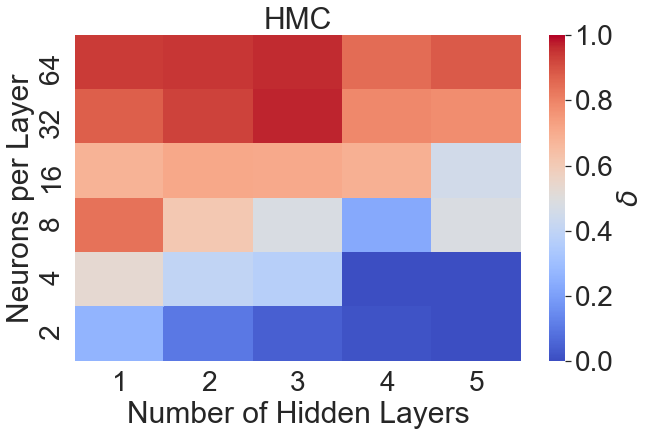}
        \includegraphics[trim={0 0 3.7cm 0},clip,width=0.29\columnwidth]{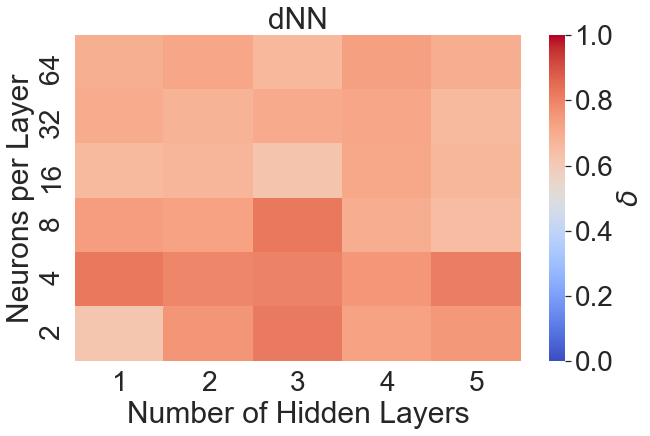}
    \includegraphics[trim={0 0 3.7cm 0},clip,width=0.29\columnwidth]{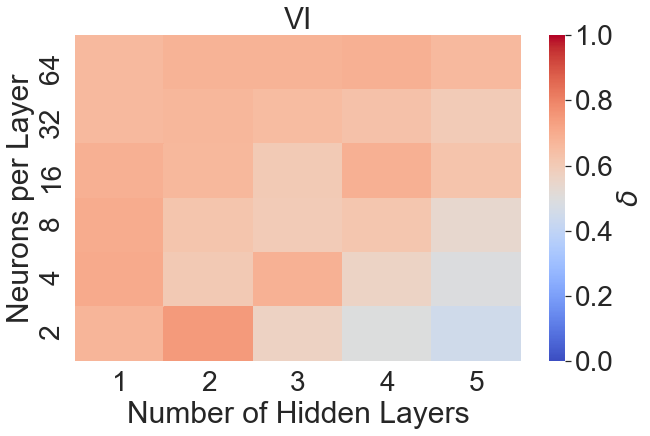}
    \includegraphics[width=0.348\columnwidth]{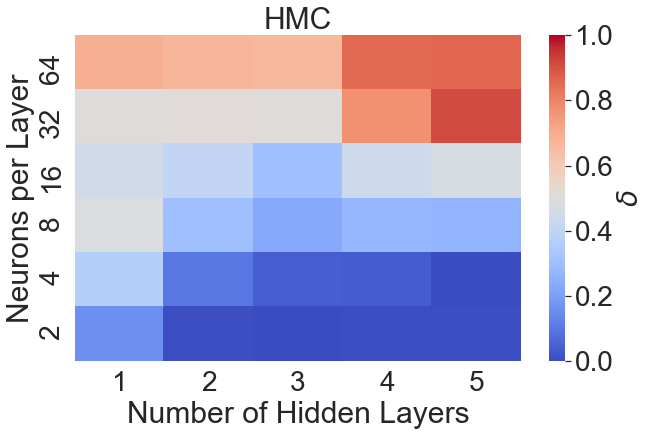}
    \caption{Heatmaps for the estimations of individual fairness ($\delta$) across various architectures, inference methods (dNN, VI or HMC), datasets, and similarity metrics. The first two rows give results for the Adult dataset (respectively \textit{Fair-FGSM} with $\ell_p$, and \textit{Fair-PGD} with Mahalanobis). The last two for the Folktables dataset.}
    \label{fig:heatmaps}
\end{figure}
We employ \textit{Fair-FGSM} and \textit{Fair-PGD} to evaluate and compare the IF of deterministic Neural Networks (dNNs) and Bayesian Neural Networks (BNNs).
We approximate the BNNs' posterior either with VI (in particular using variational online Gauss-Newton \citep{khan2018fast}) or with Hamiltonian Monte Carlo (HMC) \citep{neal2012bayesian}.
The NNs are trained to solve the taks associated to the Adult \citep{UCIdatsets} and the Folktables \citep{ding2021retiring} datasets, where we take gender as the sensitive.\footnote{Code to run the experiments can be found at \url{https://github.com/alicedoherty/bayesian-fairness}.}
%
%
%
\paragraph{Parametric Analysis}
We evaluate how IF is affected by architectural parameters.
Namely, across the two similarity metrics, we begin by fixing $\epsilon=0.1$ and analysing how $\delta$ is affected by the numbers of hidden layers and of neurons per layer, in ranges typically used for these datasets \citep{yurochkin2019training,john2020verifying,yeom2020individual,benussi2022individual}. The results for these analyses are plotted in the heatmaps in Figure \ref{fig:heatmaps}.
The three columns of heatmaps respectively show the IF estimations for models learned with dNN, VI, and HMC. 
The first two rows give the results for the Adult dataset, respectively for \textit{Fair-FGSM} with the weighted $\ell_p$ metric and \textit{Fair-PGD} with the Mahalanobis similarity metric. The last two rows show the analogous results on the Folks dataset.

We immediately observe how, in the overwhelming majority of the cases, the models learnt by HMC are markedly fairer (i.e., they have lower values of $\delta$) than their deterministic counter-parts.
The same generally applies also to VI models, except for the IF obtained on the Adult dataset by means of \textit{Fair-PGD} for the Mahalanobis metric (second row, second column).
This combined observations almost perfectly mirrors the behaviour of approximate Bayesian inference in adversarial robustness settings \citep{carbone2020robustness}, albeit here over the whole input space, and around $d_{fair}$-neighbourhoods. 
Interestingly, both adversarial robustness and overfitting generally worsen as the model becomes deeper. Instead, we observe that more often than not this pattern is reversed in IF. We posit that while the NNs are indeed becoming more fragile to attacks as they get deeper, it is also less reliant on each specific input feature, rather it builds on a non-trivial representation of them – therefore having a tendency to perform better on IF tasks. On the other hand, as expected, when the number of neurons per layer increases, we get less fair models.
%

\begin{figure}
    \centering
    \includegraphics[width=0.44\columnwidth]{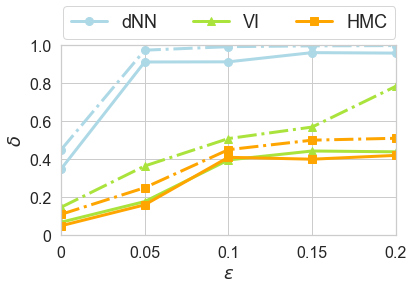}
    \includegraphics[width=0.44\columnwidth]{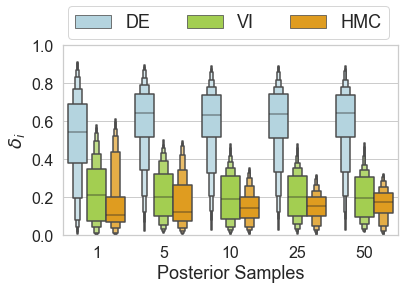}
    \caption{\textbf{Left:} Fairness for different maximum distances $\epsilon$ for deterministic and Bayesian NNs with 3 layers and either 16 (thick lines) or 32 (dashed lines) neurons per layer. \textbf{Right:} Changes in empirical distribution of fairness across deterministic and Bayesian predictors computed over an increasing number of posterior samples.}
    \label{fig:1dplots}
\end{figure}

In the left plot of Figure \ref{fig:1dplots} we analyse how individual fairness is affected by $\epsilon$, i.e., the maximum dissimilarity allowed between two individuals. We give a selection of the results for an architecture with 3 layers and 16 neurons (thick line) or 32 neurons (dashed line) trained on the Adult dataset. Of course, we observe that as $\epsilon$ increases $\delta$ increases as well, as a greater $\epsilon$ implies that we are looking at the models' behaviour on increasingly more dissimilar pairs of individuals. 
Interestingly, we find that BNNs' tendency to be more fair is consistent across the different values of epsilon.

\paragraph{Analysis of Posterior Predictive}
We experimentally inquire on the reason behind BNNs' improved individual fairness.
To this end, for a given BNN posterior predictive distribution, we solve the inner problem of Section \ref{sec:attacks} over 100 randomly sampled test points.
Rather than doing this over the full posterior, we randomly sample 15 times a number $k$ of weights realisation from the posterior, and compute the predictive posterior over them. 
We investigate values of $k$ ranging from $1$ (i.e., a single deterministic-like network extracted from the BNN) to $50$. 
We also compare BNNsr, with an averaging of deterministic neural networks in the form of a Deep Ensemble (DE) \citep{lakshminarayanan2017simple}. 
The results for this analysis are given in the right plot of Figure \ref{fig:1dplots}.
Surprisingly, even when using only one posterior sample from the BNN distribution, the models learnt with Bayesian approximation method are already significantly more fair than with the frequentist DE technique.
Furthermore, especially in the case of HMC, the more samples are drawn from the posterior distribution at prediction time, the fairer the model becomes – with a $50\%$ bias reduction when going from $1$ to $50$.
On the other hand, notice that averaging  does not help with DE, whose fairness estimation remains almost constant throughout the plot.

\section{Conclusion}
In this paper we extended adversarial attack techniques to study individual fairness (IF) for BNNs. On a set of experiments we empirically showed that BNNs tend to be intrinsically fairer than their deterministic counter-parts.  Furthermore, we showed how the particular approximate inference method has an impact on IF, with more approximate methods being less fair compared to approaches such as HMC. 
Finally, we have empirically shown how increased fairness is likely due to a combination of Bayesian training and Bayesian averaging, which may have a beneficial effect in reducing the magnitude of the gradients. 

\bibliography{jmlr-sample}


\end{document}